\documentclass[10pt]{cai}
\usepackage{amssymb}    % For \checkmark
\usepackage{tabularx}
\begin{document}
% Editorial staff will replace the following values:
% 1. Conference Year
% 2. Issue number
% 3. Article DOI
\def\conferenceyear{2026}
\volumeheader{37}{0}%{00.000}
\begin{center}

\title{Adaptive Edge Learning for Density-Aware Graph Generation
}
\maketitle

\thispagestyle{empty}

% Add Authors and Affiliations in the camera ready
% for the double blind review, please leave this section as is 
\begin{tabular}{cc}
Seyedeh Ava Razi Razavi\upstairs{\affilone}, James Sargant\upstairs{\affilone}, Sheridan Houghten\upstairs{\affilone}, Renata Dividino\upstairs{\affilone, *}
\\[0.25ex]
{\small \upstairs{\affilone} Brock University, St Catharines, Ontario, Canada.} \\
\end{tabular}

% Replace with corresponding author email address
\emails{
  \upstairs{*}rdividino@brocku.ca 
}
\vspace*{0.2in}
\end{center}

\begin{abstract}

Generating realistic graph-structured data is challenging due to discrete structures, variable sizes, and class-specific connectivity patterns that resist conventional generative modeling. While recent graph generation methods employ generative adversarial network (GAN) frameworks to handle permutation invariance and irregular topologies, they typically rely on random edge sampling with fixed probabilities, limiting their capacity to capture complex structural dependencies between nodes. We propose a density-aware conditional graph generation framework using Wasserstein GANs (WGAN) that replaces random sampling with a learnable distance-based edge predictor. Our approach embeds nodes into a latent space where proximity correlates with edge likelihood, enabling the generator to learn meaningful connectivity patterns. A differentiable edge predictor determines pairwise relationships directly from node embeddings, while a density-aware selection mechanism adaptively controls edge density to match class-specific sparsity distributions observed in real graphs. We train the model using a WGAN with gradient penalty, employing a GCN-based critic to ensure generated graphs exhibit realistic topology and align with target class distributions. Experiments on benchmark datasets demonstrate that our method produces graphs with superior structural coherence and class-consistent connectivity compared to existing baselines. The learned edge predictor captures complex relational patterns beyond simple heuristics, generating graphs whose density and topology closely match real structural distributions. Our results show improved training stability and controllable synthesis, making the framework effective for realistic graph generation and data augmentation. Source code is publicly available at~\url{https://github.com/ava-12/Density_Aware_WGAN.git}.
\end{abstract}

% add your keywords
\copyrightnotice

\section{Introduction}

Graph-structured data appears in numerous domains, from social networks and molecular structures to biological systems and communication networks, where entities and their relationships encode critical information about underlying processes and properties. The ability to generate realistic synthetic graphs has become increasingly important for applications including privacy-preserving data sharing, and data augmentation for graph-based machine learning models. However, graph generation presents fundamental challenges that distinguish it from traditional data synthesis in Euclidean spaces. Unlike images or text that have regular grid structures and fixed dimensionality, graphs exhibit irregular topologies, variable sizes, and lack canonical node ordering. Successful graph generators must capture complex dependencies between topology and node attributes while maintaining the structural properties that characterize different graph types.

Early deep learning approaches to graph generation have explored various strategies to address these challenges. Methods such as DeepGMG~\cite{li2018learning} and GraphRNN~\cite{you2018graph} model graphs through sequential generation processes, while GraphVAE~\cite{simonovsky2018graphvae} leverages variational autoencoders to learn latent graph representations. Generative Adversarial Networks (GANs) have shown particular promise, with GraphGAN~\cite{wang2018graphgan}, EGraphGAN~\cite{wang2024graph}, and MolGAN~\cite{de2018molgan} demonstrating the potential of adversarial training for graph synthesis. LGGAN~\cite{fan2019labeled} further advanced the field by introducing labeled graph generation with comprehensive evaluation metrics including Maximum Mean Discrepancy (MMD) over degree, clustering, and orbit statistics. However, these GAN-based approaches often suffer from training instability, mode collapse, and difficulties in generating graphs with variable sizes.

The introduction of Wasserstein GANs (WGANs)~\cite{arjovsky2017wasserstein} marked a significant advancement in addressing the instability issues inherent in standard GANs. WGAN-GP~\cite{gulrajani2017improved} further improved upon this by enforcing the Lipschitz constraint through gradient penalty rather than weight clipping, enabling more effective training and better sample diversity. Recent work has successfully adapted the WGAN framework to graph generation by combining it with Graph Neural Networks (GNNs) as critics. In ~\cite{ava_cascon}, a generator produces both node features and connectivity patterns, while a GNN-based critic evaluates graph realism and class consistency. This method demonstrated superior performance in generating class-conditional graphs with improved training stability. However, despite these advances, the approach relies on simplistic edge generation using fixed, class-dependent probabilities and random sampling. This limitation results in graphs with unrealistic connectivity patterns that fail to capture the complex structural dependencies between nodes and do not reflect the learned node feature distributions.

We address these core limitations by developing a new WGAN-based framework featuring a graph generator that employs a learnable, distance-driven edge predictor, enabling the model to infer connectivity patterns and capture complex structural dependencies. We further introduce a density-aware edge selection mechanism that computes probabilities for all possible node pairs and selects edges based on class-specific density statistics, ensuring generated graphs maintain realistic sparsity patterns. Additionally, our architecture employs node-specific latent vectors rather than shared graph-level noise, promoting diversity in node features within each graph. Experimental results show improved graph generation that matches the statistical properties of the training distribution. The generated graphs exhibit coherent structural patterns, where connectivity meaningfully reflects learned relationships between node features.

\section{Foundations}
\label{sec:background}
We provide the formal definitions and terminology used in this work. Let a graph be represented as $G=(V,E,X,A,y)$ with a set of nodes $V = \{v_1,...,v_N\}$ and edges $E \subseteq V \times V$. $X \in \mathbb{R}^{N \times d}$ is the node feature matrix where $X_i \in \mathbb{R}^d$ corresponds to the $d$-dimensional feature vector of node $v_i \in V$. The adjacency matrix $A \in \{0,1\}^{N\times N}$ encodes the graph structure, where $A_{ij} = 1$ if edge $(v_i, v_j) \in E$ and $0$ otherwise. For graph-level tasks, $y \in \{1, ..., C\}$ denotes the graph class label from $C$ distinct classes.

\subsection{Graph Neural Networks}

Graph Neural Networks (GNNs) have enabled the training of deep learning models on graph-structured data, aiming to represent graphs as low-dimensional embeddings that can be leveraged for downstream tasks such as node classification, link prediction, and graph classification. A number of architectures have been proposed~\cite{hamilton2017inductive, kipf2017semi, velivckovic2018graph, xu2018powerful}, following the message-passing scheme. GNNs learn a multi-step mapping: $f : G \rightarrow Z \in \mathbb{R}^{N \times d'}$, where $Z$ is the updated feature matrix for a graph $G$, with $d'$ being the output embedding dimension, and each row of $Z$ represents a low-dimensional embedding of a node in $G$.

Each GNN layer typically consists of two phases: \textit{Aggregation}, where information from the neighborhood of each node is gathered using a permutation-invariant function, and \textit{Update}, where each node's feature vector is updated based on its current state and the aggregated neighborhood information. These steps are applied recursively over $k$ layers in the GNN. The hidden representation $\mathbf{h}_v^{(l)}$ of a node $v$ at layer $l$ is computed as:
\begin{equation}\label{eq:gnn_update}
\mathbf{h}_v^{(l)} = \sigma \left( \mathbf{B}_l \mathbf{h}_v^{(l-1)} + \mathbf{W}_l \sum_{u \in \mathcal{N}g(v)} g(\mathbf{h}_u^{(l-1)}) \right), 
\end{equation}
where $\mathbf{h}_v^{(0)} = X_v$ is the initial feature vector of node $v$, $\mathcal{N}g(v)$ is the set of neighbors of $v$, $\mathbf{B}_l$ and $\mathbf{W}_l$ are learnable weight matrices at layer $l$, $g$ is a differentiable, permutation-invariant aggregation function (e.g., mean, sum, or max), and $\sigma$ is a nonlinear activation function.

The features of node $v$ and its neighbors are combined by an update function, such as multi-layer perceptron (MLP), to produce new node embeddings. The output of the final layer, denoted as $Z^{(k)}$, can then be used for various downstream tasks.

\subsection{Generative Adversarial Networks}

Generative Adversarial Networks (GANs) constitute a powerful framework for learning complex data distributions through an adversarial training process~\cite{goodfellow2014generative}. The core principle involves two neural networks competing in a minimax game: a generator network $G$ that learns to produce synthetic data samples, and a discriminator network $D$ that learns to distinguish between real and generated samples. The generator maps random noise vectors $z \sim p_z(z)$ from a prior distribution to the data space, while the discriminator outputs a probability indicating whether its input is real or synthetic.

The training objective is formulated as a minimax optimization problem:
\begin{equation}
\min_G \max_D V(D, G)
= \mathbb{E}_{x \sim \mathbb{P}_r}[\log D(x)]
+ \mathbb{E}_{z \sim p_z}[\log(1 - D(G(z)))].
\end{equation}
During training, the discriminator maximizes its ability to distinguish real and fake samples, while the generator minimizes this ability.

\subsubsection{Wasserstein GANs}

Standard GANs often suffer from training instability, mode collapse, and vanishing gradients. Wasserstein GANs (WGANs)~\cite{arjovsky2017wasserstein} address these issues by replacing the Jensen--Shannon divergence with the Wasserstein-1 (Earth-Mover) distance, which provides smoother gradients and a more meaningful loss metric. The Wasserstein distance between the real and generated distributions $\mathbb{P}_r$ and $\mathbb{P}_g$ is defined as:
\begin{equation}
    W(\mathbb{P}_r,\mathbb{P}_g)
    = \inf_{\gamma \in \Pi(\mathbb{P}_r,\mathbb{P}_g)}
    \mathbb{E}_{(x,x')\sim\gamma} \big[ \|x - x'\| \big],
\end{equation}
where $\Pi(\mathbb{P}_r,\mathbb{P}_g)$ denotes the set of all couplings with marginals $\mathbb{P}_r$ and $\mathbb{P}_g$. Using the Kantorovich--Rubinstein dual formulation, the WGAN objective becomes:
\begin{equation}
    \min_G \max_{D \in \mathcal{D}}
    \mathbb{E}_{x \sim \mathbb{P}_r}[D(x)]
    - \mathbb{E}_{z \sim p_z}[D(G(z))],
\end{equation}
where $\mathcal{D}$ is the set of 1-Lipschitz critics. To enforce this constraint, WGAN-GP~\cite{gulrajani2017improved} introduces a gradient penalty:
\begin{equation} \label{eq:penalty}
    \mathcal{L}_{\mathrm{GP}}
    = \lambda \,
    \mathbb{E}_{\tilde{x}}
    \big[ \left( \|\nabla_{\tilde{x}} D(\tilde{x})\|_2 - 1 \right)^2 \big],
\end{equation}
where $\tilde{x} = \epsilon x_{\mathrm{real}} + (1-\epsilon)x_{\mathrm{fake}}$ with $\epsilon \sim U[0,1]$ representing interpolations between real and generated samples.

\section{Related Work}

Graph generation models have played a significant role in biology, social sciences, and applications such as data augmentation, drug discovery, and anomaly detection since their introduction by Erdős and Rényi~\cite{erdos1959random}. Probabilistic block models represent an important class of graph generators. The Stochastic Block Model (SBM)~\cite{goodfellow2016nips} assigns nodes to groups, with connection probability depending on group membership. The Mixed Membership Stochastic Blockmodel (MMSB) extended SBM by allowing nodes to belong to multiple groups simultaneously~\cite{airoldi2008mixed}. To address MMSB's limitations including scalability issues, fixed community numbers, and poor performance on sparse graphs, dMMSB added a state-space model for evolving mixed memberships over time~\cite{xing2010state}.

Deep learning approaches have revolutionized sequential graph generation. DeepGMG~\cite{li2018learning} formulates graph generation as a sequential decision process, adding nodes and edges one at a time using neural network probabilities. It employs GNNs with message-passing to maintain hidden representations and encode local structural information. The model supports conditional generation and handles multiple graph construction orderings through fixed or random node orderings during training. GraphRNN~\cite{you2018graph} generates graphs sequentially by modeling the graph as a sequence, and it is evaluated by comparing the generated and real graphs on properties such as degree distribution and clustering coefficients. Using two RNN components, a graph-level RNN for node addition and an edge-level RNN for connectivity prediction, GraphRNN captures both global structure and local patterns. GraphRNN uses BFS ordering to standardize graph representation and offers GraphRNN-S, a simplified variant assuming conditional edge independence for faster training. GraphVAE~\cite{simonovsky2018graphvae} pioneered variational autoencoders for graph generation, addressing the challenges of generating discrete graph structures. It learns a continuous latent space to encode and decode entire graphs, particularly for small molecular structures.

Closely related to our framework are adversarial methods for graph synthesis. GraphGAN~\cite{wang2018graphgan} combines generative and discriminative models through adversarial training, with the generator modeling neighbor distributions using graph-aware softmax over BFS trees. Despite strong downstream performance, it suffered from training instability. EGraphGAN~\cite{wang2024graph} addressed this using evolutionary algorithms with mutation strategies for sampling. For direct graph generation, MolGAN~\cite{de2018molgan} generates molecular graphs using reinforcement learning but lacks flexibility for varying structures. LGGAN~\cite{fan2019labeled} overcomes this limitation, generating diverse graph types using MLPs to produce node label matrices and adjacency matrices. These approaches, however, typically generate edges through random sampling or direct matrix prediction without modeling structural dependencies. Our method differs by learning edge formation through distance-based predictions in a latent space and incorporating explicit density-aware selection to match class-specific sparsity patterns, enabling more structured and realistic graph synthesis.

\section{Methodology}

\begin{figure}
    \centering 
    \includegraphics[width=1\linewidth]{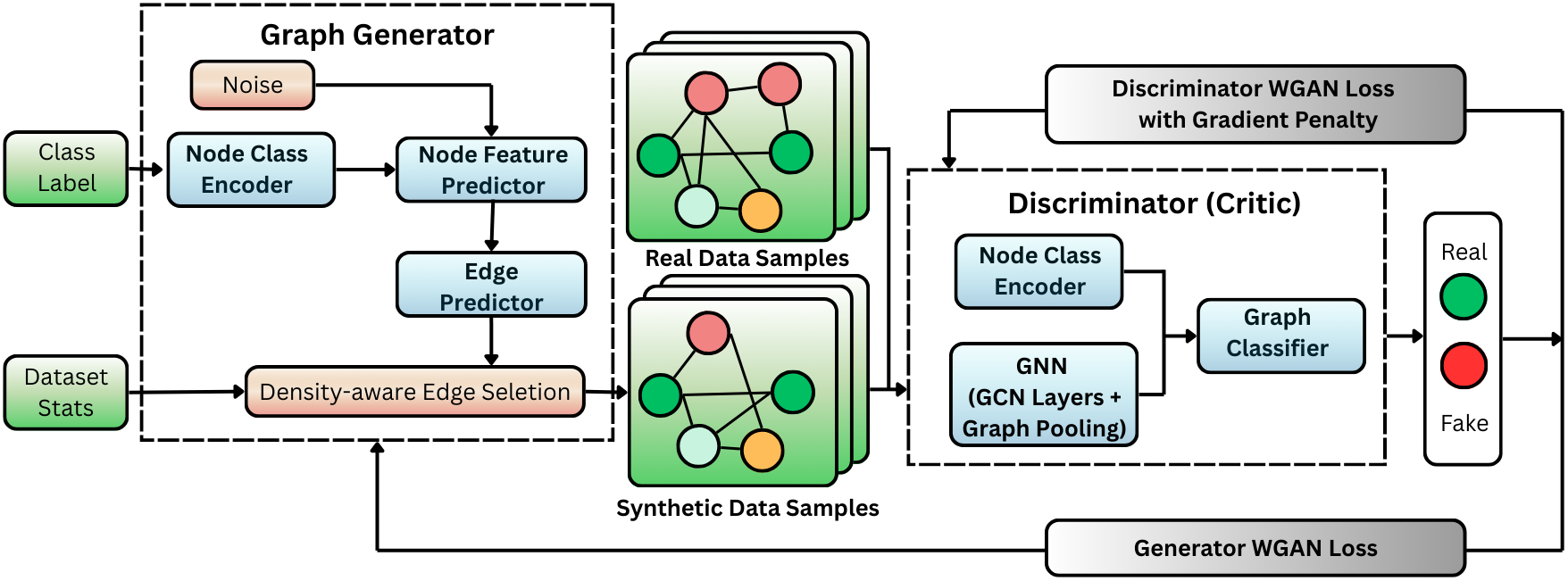}
    \caption{%Overview of our conditional graph generation framework. 
    Model overview: the generator produces diverse node features from individual noise vectors combined with class embeddings. The distance-based edge predictor learns to map nodes into a latent space where proximity determines connectivity. Edge selection follows class-specific density statistics extracted from training data. The GNN-based critic processes graphs through multiple convolution layers, pools node features, and combines them with class embeddings to produce Wasserstein scores that ensure both distributional similarity and class alignment. Note that blue represents learning modules. In this approach, the probability distribution remains fully differentiable, allowing the predictor to learn directly from the WGAN objective.}
    \label{fig:arch}
\end{figure}

Our objective is to generate class-conditional synthetic graphs that mirror the structural characteristics of real data. Let $\mathcal{G}=\{(G_i,y_i)\}_{i=1}^N$ denote a set of labeled training graphs, where each $G_i=(V_i,E_i,X_i)$ contains nodes, edges, and node features, and $y_i \in \{1,\dots,C\}$ is a graph label. We learn a generator $G_\theta:\mathcal{Z}\times\mathcal{Y}\rightarrow \mathcal{G}$ that maps a latent input and a class condition to a graph whose distribution approximates $\mathbb{P}_r$. Unlike prior GAN-based approaches that rely on fixed-probability random edge sampling~\cite{ava_cascon,fan2019labeled}, our method learns connectivity patterns directly from data. This change gives the model the ability to infer which nodes are likely to connect, rather than deciding edges from fixed probabilities. As shown in Figure~\ref{fig:arch}, our architecture consists of a graph generator with a distance-based edge predictor and density-aware edge selection, and a discriminator for adversarial training.

\subsection{Graph Generator}

The generator creates graphs based on class-specific patterns through three stages: (1) class conditioning provides structural guidance for the target graph type; (2) node feature synthesis generates diverse node attributes appropriate for the class; and (3) edge construction learns connectivity patterns by representing structural relationships in a latent space.

\begin{description}[leftmargin=0cm]
    \item[Class Encoder] Class information is encoded through an embedding layer that maps class labels to dense vectors $e_y$. This embedding captures class-specific structural properties and is shared across all nodes in a generated graph to ensure consistent structure.% within the target class.
    
    \item[Node Feature Predictor] To generate diverse graphs within the same class, each node must have both shared class characteristics and individual variation. We provide each node with an independent noise vector $z_i \sim \mathcal{N}(0, I)$ to introduce necessary randomness for generating different structures. The generator combines this noise with the class embedding to produce node features $x_i = \textrm{MLP}_{\mathrm{node}}([z_i; e_y])$, creating a feature matrix $X \in \mathbb{R}^{n \times d}$ that balances class consistency with structural diversity. 
    
    Different graph classes have different typical sizes. For example, molecular graphs usually contain 10-50 atoms while social networks may have thousands of nodes. We preserve this variation by estimating the size distribution for each class and sampling $n \sim \mathrm{Clip}(\mathcal{N}(\mu_c, cf\sigma_c), n^c_{\min}, n^c_{\max})$, where $\mu_c$ and $\sigma_c$ represent the typical size and variation for class $c$. The factor $cf$ prevents unrealistic extreme sizes while maintaining natural variation, and clipping ensures generated sizes stay within observed ranges.

    \item[Edge Predictor] Random edge sampling with fixed probabilities cannot capture the structured patterns in real graphs. In real networks, nodes with similar roles or attributes tend to connect more often. We introduce a learnable edge predictor that estimates connection likelihood based on learned structural similarity (i.e., graph homophily). This design follows the principle that similar nodes are more likely to connect, while remaining flexible enough to capture other patterns such as hub structures. The predictor maps nodes into a latent space where closeness reflects connection likelihood. For each node pair $(i,j)$, we compute the edge probability as:
    \begin{equation}
    p_{ij} = \sigma\left(\frac{-\|h_i - h_j\|_2 + \theta}{T}\right),
    \end{equation}
    where the distance $\|h_i - h_j\|_2$ measures how different the node representations are. The learnable threshold $\theta$ determines the critical distance for edge formation and adapts to each class during training. The temperature $T$ controls how sharp the decision boundary is, allowing smooth gradients during training. This approach enables the model to learn meaningful connection patterns rather than using fixed random rules.

    \item[Density-Aware Edge Selection] Real graphs have very different density patterns depending on their domain. Molecular graphs are sparse and locally structured due to chemical constraints, while social networks may be globally sparse but contain dense local groups. Capturing these class-specific density patterns is essential for generating realistic graphs.
    
    We compute the expected edge density for each class:
    \begin{equation}
    \rho_c = \frac{2\bar{m}_c}{\bar{n}_c(\bar{n}_c - 1)}
    \end{equation}
    where $\bar{m}_c$ and $\bar{n}_c$ are the average numbers of edges and nodes for class $c$. This density $\rho_c$ captures the typical sparsity level for each graph type.
    
    During generation, after computing edge probabilities for all $\binom{n}{2}$ possible pairs, we select the top $k = \lfloor \rho_c \cdot \binom{n}{2} \rfloor$ edges with highest probabilities. This selection serves three purposes: (1) matching the typical density of the target class, (2) using the model's most confident predictions, and (3) guiding the edge predictor to learn realistic patterns. For undirected graphs, each selected edge is added in both directions to form the final edge set $E$.

\end{description}

\subsection{Discriminator (Critic)}

The discriminator evaluates whether generated graphs have the structural patterns found in real graphs from the target class.

\begin{description}[leftmargin=0cm]

    \item[Graph Encoder] 
    The discriminator must evaluate structure at multiple levels: local patterns (triangles, stars), intermediate community structure, and global properties (connectivity, diameter). We use a GNN that builds representations through neighborhood aggregation (Equation~\ref{eq:gnn_update}). Over $L$ layers, each node incorporates information from increasingly distant neighbors. This allows the discriminator to check whether generated graphs have consistent structural patterns at all scales. The resulting node embeddings $Z^{(L)} \in \mathbb{R}^{n \times d'}$ encode structural information. Global pooling creates a graph-level representation that is independent of node ordering, allowing the discriminator to evaluate graphs of different sizes consistently.
    \item[Class Encoder] 
    To evaluate class-specific realism, the discriminator must compare generated structures against the expected patterns for that class. We combine the graph representation with the class embedding $e_y$ to form $f = [g; e_y] \in \mathbb{R}^{d' + d_{\text{class}}}$, allowing the critic to assess whether the structure matches what is typical for the target class.
    \item[Graph Classifier] 
    The combined representation is processed by an MLP critic that outputs a Wasserstein score $D_\phi(G, y) = \text{MLP}_\phi(f)$. This critic produces continuous scores that reflect structural quality: higher values indicate graphs whose patterns closely match real samples from class $y$, while lower values indicate structural problems. This scoring provides informative gradients that guide the generator toward more realistic synthesis.

\end{description}

\subsection{Training Procedure}

Training follows the WGAN-GP framework described above. In our implementation, we use a \emph{conditional critic} $D_{\phi}(x,y)$, where $y$ denotes the conditioning information.

The critic is trained to minimize:
\begin{equation}
\mathcal{L}_D
= - \mathbb{E}_{x \sim \mathbb{P}_r}[D_{\phi}(x, y)]
+ \mathbb{E}_{\hat{x} \sim \mathbb{P}_g}[D_{\phi}(\hat{x}, y)]
+ \lambda \,\mathcal{L}_{\mathrm{GP}},
\end{equation}
where $\hat{x} = G(z,y)$ and $\mathbb{P}_g$ is the model distribution induced by the generator. The gradient penalty term reuses the definition in Equation~\ref{eq:penalty}, adapted to the conditional critic:
\begin{equation}
\mathcal{L}_{\mathrm{GP}}
= \mathbb{E}_{\tilde{x}}
\left[
\left(
\|\nabla_{\tilde{x}} D_{\phi}(\tilde{x}, y)\|_2 - 1
\right)^2
\right],
\end{equation}
where $\tilde{x}$ denotes interpolations between real and generated samples.

To stabilize edge formation during training, we apply temperature annealing to the edge predictor:
\begin{equation}
T(t) = \max(T_{\text{end}}, T_{\text{start}} - \alpha \cdot t)
\end{equation}

The temperature parameter $T$ governs the sharpness of the edge probability function. High initial temperature produces soft probability outputs, encouraging exploration of diverse graph structures and preventing collapse into limited edge patterns. As training progresses, $T$ is gradually reduced, increasing contrast between close and distant node pairs and causing the edge predictor to shift from probabilistic sampling to confident, near-binary decisions. This annealing schedule provides smooth transition from stochastic graph formation early in training to sharper, more realistic structures later. This helps the generator converge to stable and expressive graph distributions while maintaining diversity during learning.

\section{Evaluation}
\label{sec:exp}

In this section, we empirically evaluate our graph generation framework’s ability to learn and reproduce complex graph distributions across diverse domains. We assess generation quality using a multi-faceted protocol that measures distributional fidelity, diversity, and structural preservation. Our experiments address three key questions: (i) how accurately the model captures the statistical properties of real graphs, (ii) whether it can generate diverse and novel graphs while preserving class-specific characteristics, and (iii) how it compares to existing graph generation methods. Source code is publicly available at~\url{https://github.com/ava-12/Density_Aware_WGAN.git}.

\subsection{Experimental Setup}

%\subsubsection{Datasets} 
\begin{description}[leftmargin=0cm]
\item[Datasets]  We evaluate our method on three benchmark graph classification datasets~\cite{morris2020tudataset} that span different domains and structural complexities as shown in the table below. MUTAG(188 graphs) represents chemical compounds for mutagenicity prediction, with small graphs averaging 18 nodes. ENZYMES (600 graphs) contains protein structures across 6 enzyme classes, with moderate-sized graphs of approximately 33 nodes. PROTEINS (1,113 graphs) provides the largest and most structurally diverse dataset for enzyme/non-enzyme classification, with graphs ranging from 4 to 620 nodes. These datasets vary significantly in their feature representations: MUTAG includes both 7-dimensional node and 4-dimensional edge features, while ENZYMES and PROTEINS utilize only 3-dimensional node features, providing a comprehensive testbed for evaluating graph generation performance across different scales and complexities.

%\begin{table}[htbp]
%\centering
\begin{center}
\begin{tabular}{lccc}
\toprule
\textbf{Property} & \textbf{MUTAG} & \textbf{ENZYMES} & \textbf{PROTEINS} \\
\midrule
\textbf{Domain} & Chemistry & Biochemistry & Biochemistry \\
\textbf{Task Type} & Binary & Multi-class (6) & Binary \\
\textbf{Number of Graphs} & 188 & 600 & 1,113 \\
\textbf{Number of Classes} & 2 & 6 & 2 \\
\textbf{Avg. Nodes} & 17.93 & 32.63 & 39.06 \\
\textbf{Avg. Edges} & 19.79 & 62.14 & 72.82 \\
\textbf{Node Features} & 7 & 3 & 3 \\
%\textbf{Edge Features} & 4 & None & None \\
\bottomrule
\end{tabular}
\end{center}
%\label{tab:dataset_characteristics}
%\end{table}

%\subsubsection{Evaluation Metrics}
\item[Evaluation Metrics] We evaluate generated graphs using \textit{Maximum Mean Discrepancy}(MMD)~\cite{gretton2012kernel}, which measures distribution similarity in a reproducing kernel Hilbert space:
\begin{equation}
\text{MMD}^2(P, Q) = \mathbb{E}_{x,x' \sim P}[k(x,x')] 
+ \mathbb{E}_{y,y' \sim Q}[k(y,y')] 
- 2\,\mathbb{E}_{x \sim P, y \sim Q}[k(x,y)],
\label{eq:mmd}
\end{equation}
where $k(\cdot,\cdot)$ is a positive-definite kernel, $P$ is the real distribution, and $Q$ is the generated distribution.

We compute MMD over three complementary graph statistics. \textit{Degree distribution} captures local connectivity by comparing empirical degree histograms, where $h_G[i]$ denotes the fraction of nodes with degree $i$. \textit{Clustering coefficient} measures triangle formation density through the local clustering coefficient:
\begin{equation}
C_i = \frac{2 \cdot |\{e_{jk} : v_j, v_k \in N(v_i),\, e_{jk} \in E\}|}{k_i(k_i - 1)},
\end{equation}
aggregated into histograms for nodes with degree $k_i > 1$. \textit{Spectral features} assess global topology using eigenvalues of the normalized Laplacian $L = I - D^{-1/2} A D^{-1/2}$. These three metrics provide complementary views: degree distribution captures local connectivity, clustering coefficients reveal community structure, and spectral features encode global topology. We combine them as:
\begin{equation}
    \text{MMD}_{\text{combined}} = \alpha \cdot \text{MMD}_{\text{degree}} + \beta \cdot \text{MMD}_{\text{clustering}} + \gamma \cdot \text{MMD}_{\text{spectral}},
\end{equation}
where $\alpha + \beta + \gamma = 1$ and weights are determined empirically. Additionally, we measure \textit{uniqueness} (proportion of distinct generated graphs) and \textit{novelty} (proportion not in the training set) to assess diversity and generalization beyond memorization.

%\subsubsection{Baseline Models} 
\item[Baseline Models] We compare our approach against state-of-the-art graph generation models: DeepGMG~\cite{li2018learning}, GraphRNN~\cite{you2018graph}, LGGAN~\cite{fan2019labeled}, and WPGAN~\cite{ava_cascon}.

%\subsubsection{Implementation Details}
\item[Implementation Details] Our discriminator implements a Graph Convolutional Network (GCN)~\cite{kipf2017semi} with global mean pooling, $g = \frac{1}{n} \sum_{i=1}^{n} h_i^{(L)}$, which summarizes the graph by averaging node representations after multi-hop neighborhood aggregation. While we use mean pooling in our experiments, alternative operators such as sum or max pooling can be employed. The framework is compatible with other GNN architectures, including GAT~\cite{velivckovic2018graph} and GraphSAGE~\cite{hamilton2017inductive}, which we leave for future exploration. The table below presents the architecture parameters and training configuration optimized for stable adversarial learning. The hyperparameter choices reflect established best practices for WGAN training: the 5:1 critic-to-generator training ratio ensures accurate Wasserstein distance estimation, while the 2.5:1 learning rate ratio maintains adversarial balance. The reduced momentum ($\beta_1 = 0.5$) is standard for GAN training to improve convergence stability. We employ temperature annealing for the edge predictor, gradually reducing $T$ from 2.0 to 0.5 over approximately 50 epochs to transition from exploratory edge formation to confident connectivity decisions.

\begin{center}
\begin{tabular}{llccc}
\toprule
\textbf{Component} & \textbf{Parameter} 
& \textbf{MUTAG} & \textbf{ENZYMES} & \textbf{PROTEINS} \\
\midrule
\multirow{3}{*}{Architecture} 
 & Noise dimension      & \multicolumn{3}{c}{16} \\
 & Class embedding      & \multicolumn{3}{c}{8}  \\
 & Hidden dimension     & \multicolumn{3}{c}{32} \\
\midrule
\multirow{7}{*}{Training} 
 & Critic iterations ($n_{\text{critic}}$) 
 & 5 & 5 & 5 \\
 & Gradient penalty ($\lambda_{GP}$) 
 & 10 & 10 & 10 \\
 & Temperature schedule ($T$)
 & \multicolumn{3}{c}{$2.0 \to 0.5$ (~50 epochs)} \\
 & Learning rate (Generator) 
 & $1\times10^{-4}$ & $3\times10^{-4}$ & $2\times10^{-4}$ \\
 & Learning rate (Discriminator) 
 & $5\times10^{-4}$ & $5\times10^{-4}$ & $5\times10^{-4}$ \\
 & Adam betas 
 & (0.5, 0.9) & (0.5, 0.9) & (0.5, 0.9) \\
 & Batch size 
 & 64 & 64 & 64 \\
\midrule
\multirow{3}{*}{Evaluation} 
 & $\alpha$ & \multicolumn{3}{c}{0.4} \\
 & $\beta$  & \multicolumn{3}{c}{0.4} \\
 & $\gamma$ & \multicolumn{3}{c}{0.2} \\
\bottomrule
\end{tabular}
\end{center}

%\subsection{Results and Analysis}
\item[Results and Analysis]

\begin{table}[htbp]
\centering
\caption{Detailed experimental results across all datasets and classes}
\resizebox{\textwidth}{!}{%
\begin{tabular}{llccccccccc}
\toprule
\textbf{Dataset} & \textbf{Class} & \textbf{MMD} & \textbf{MMD} & \textbf{MMD} & \textbf{Avg Nodes} & \textbf{Avg Edges} & \textbf{MMD} & \textbf{Unique-} & \textbf{Novelty} \\
 & & \textbf{Degree} & \textbf{Clustering} & \textbf{Spectral} & \textbf{(Real$\rightarrow$Gen)} & \textbf{(Real$\rightarrow$Gen)} & \textbf{Combined} & \textbf{ness} & \\
\midrule
\multirow{2}{*}{\textbf{MUTAG}} 
 & Class 0 & 0.361 & 1.082 & 0.12 & 13.4$\rightarrow$13.4 & 14.0$\rightarrow$13.7 & 0.577 & 0.958 & 0.912 \\
 & Class 1 & 0.249 & 1.04& 0.105 & 20.7$\rightarrow$19.7 & 23.5$\rightarrow$22.1 & 0.522 & 0.974 & 0.975  \\
\midrule
\multirow{6}{*}{\textbf{ENZYMES}} 
 & Class 0 & 0.158 & 0.142 & 0.095 & 31.6$\rightarrow$41.0 & 63.3$\rightarrow$86.0 & 0.133 & 0.992 & 0.986 \\
 & Class 1 & 0.153 & 0.143 & 0.08 & 32.1$\rightarrow$31.6 & 62.1$\rightarrow$68.9 & 0.127 & 0.992 & 0.931 \\
 & Class 2 & 0.15 & 0.142 & 0.072 & 30.1$\rightarrow$30.5 & 60.3$\rightarrow$62.2 & 0.124 & 0.988 & 0.917 \\
 & Class 3 & 0.170 & 0.149 & 0.081 & 37.4$\rightarrow$37.1 & 73.5$\rightarrow$71.2 & 0.135 & 1.000 & 0.962 \\
 & Class 4 & 0.141 & 0.129 & 0.098 & 32.1$\rightarrow$28.4 & 60.4$\rightarrow$53.0 & 0.122 & 1.000 & 0.982\\
 & Class 5 & 0.214 & 0.173 & 0.08 & 38.8$\rightarrow$27.7 & 77.3$\rightarrow$54.0 & 0.158 & 0.979 & 0.953 \\
\midrule
\multirow{2}{*}{\textbf{PROTEINS}} 
 & Class 0 & 0.063 & 0.042 & 0.146 & 54.5$\rightarrow$49.5 & 103.0$\rightarrow$124.9 & 0.08 & 0.955 & 0.994 \\
 & Class 1 & 0.082 & 0.055 & 0.056 & 19.6$\rightarrow$25.1 & 36.5$\rightarrow$57.2 & 0.063 & 0.913& 0.838 \\
\bottomrule
\end{tabular}
}
\label{tab:results}
\end{table}

\begin{table*}[t!]
\centering
\caption{Comparison with state-of-the-art methods using MMD metrics (lower is better) as presented in~\cite{fan2019labeled,ava_cascon}. Best results are bolded.}
\label{tab:mmd_comparison}
\begin{tabular}{lcccccc}
\toprule
\textbf{Model} & \multicolumn{3}{c}{\textbf{PROTEINS}} & \multicolumn{3}{c}{\textbf{ENZYMES}} \\
 & Degree & Clustering & Spectral & Degree & Clustering & Spectral\\
\midrule
DeepGMG~\cite{li2018learning} & 0.96 & 0.63 & -- & 0.43 & 0.38 & -- \\
GraphRNN~\cite{you2018graph} & 0.04 & 0.18 & -- & 0.06 & 0.20 & -- \\
LGGAN~\cite{fan2019labeled} & 0.18 & 0.15 & -- & 0.09 & 0.17 & -- \\
WPGAN~\cite{ava_cascon} & \textbf{0.03} & 0.31 & -- & \textbf{0.02} & 0.28 & -- \\
\midrule
\textbf{Our Approach} & 0.09 & \textbf{0.07} & 0.07 & 0.10 & \textbf{0.08} & 0.05  \\
\bottomrule
\end{tabular}
\end{table*}

Table~\ref{tab:results} presents a detailed breakdown of performance across all datasets and classes. For the ENZYMES dataset, the model achieves strong and consistent performance across all six classes, with combined MMD values ranging from 0.106 to 0.182. The model demonstrates particularly strong spectral alignment (MMD values between 0.042-0.133) and clustering preservation (0.114-0.200), indicating effective capture of both local motifs and global topology. Degree distribution alignment is also strong, with MMD values between 0.137-0.208. The model achieves near-perfect uniqueness scores (0.941-1.000) and high novelty (0.833-1.000) across all classes, confirming diverse generation without memorization.

On the PROTEINS dataset, the model achieves the lowest combined MMD scores overall (0.066-0.089), demonstrating excellent structural fidelity. Performance is particularly strong on clustering coefficients (0.039-0.050) and degree distributions (0.054-0.091), with Class 1 showing the best overall alignment (combined MMD of 0.066). Spectral MMD values are slightly higher for Class 0 (0.192) compared to Class 1 (0.062), reflecting differences in graph sizes between classes. High uniqueness (0.921-0.951) and novelty (0.859-1.000) scores confirm the model generates diverse, non-memorized samples.

On the MUTAG dataset, the model accurately reproduces graph size statistics (average nodes and edges closely match real distributions) while maintaining high diversity (uniqueness 0.857-0.933, novelty 0.933-1.000). However, clustering coefficient MMD values are notably higher (1.053--1.059) compared to other datasets. This is attributable to MUTAG's small graph sizes (13-21 nodes) and limited structural diversity, which makes distributional matching more challenging. Despite this, combined MMD scores (0.527-0.536) indicate reasonable overall structural alignment, and low degree (0.249--0.266) and spectral (0.102-0.107) MMD values demonstrate the model still captures fundamental connectivity patterns.

Table~\ref{tab:mmd_comparison} compares our approach against state-of-the-art methods on PROTEINS and ENZYMES datasets. On PROTEINS, our model achieves the best clustering coefficient preservation (MMD of 0.07) among all methods, significantly outperforming LGGAN (0.15), GraphRNN (0.18), and WPGAN (0.31), demonstrating superior modeling of local connectivity and community structure. For degree distribution, our approach achieves competitive performance (0.08) compared to WPGAN's best result (0.03). We also report spectral MMD (0.06), which captures global topology, a metric not evaluated by prior work.  

\begin{figure}[t!]
    \centering
    \includegraphics[width=\linewidth]{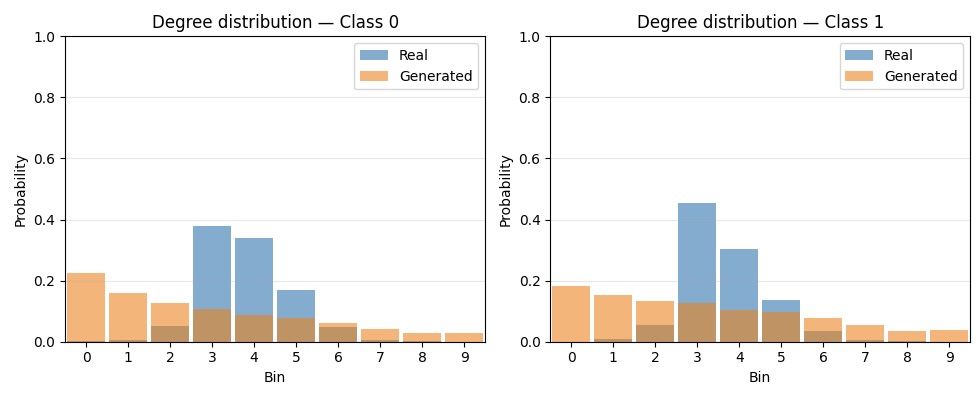}
    \vspace{0.5em}
    \includegraphics[width=\linewidth]{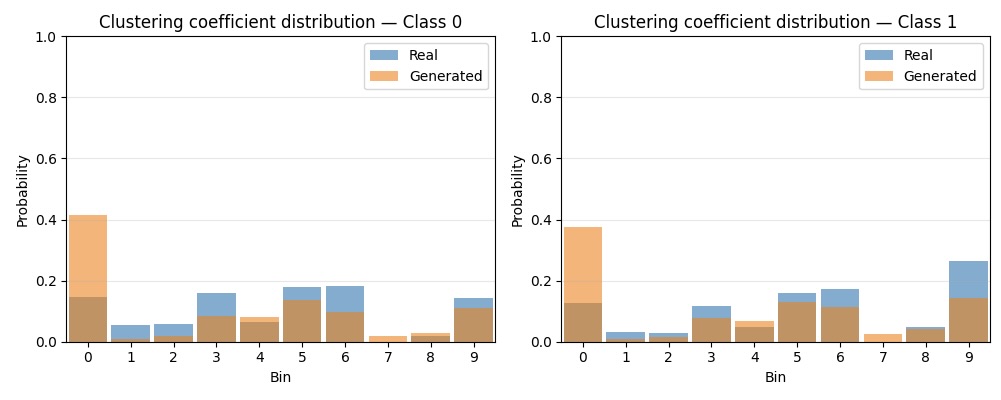}
    \caption{Distribution comparison between real and generated graphs on the PROTEINS dataset. Top: Degree distribution shows the model concentrates predictions on mid-range values (bins 2--6), producing narrower variance compared to real graphs. This reflects our deterministic density-aware edge selection, which enforces class-average sparsity but may over-regularize individual degree heterogeneity. Bottom: Clustering coefficient distribution demonstrates strong alignment with substantial overlap, confirming the model's superior preservation of local connectivity patterns and community structure, consistent with the best clustering MMD (0.07) achieved among compared methods.}
    \label{fig:proteindist}
\end{figure}

\begin{figure}[t!]
    \centering
    \begin{minipage}{0.24\textwidth}
        \centering
        \includegraphics[width=\linewidth]{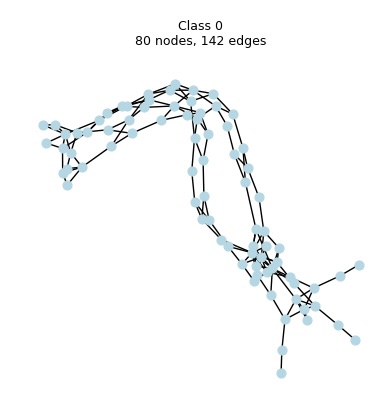}
    \end{minipage}\hfill
    \begin{minipage}{0.24\textwidth}
        \centering
        \includegraphics[width=\linewidth]{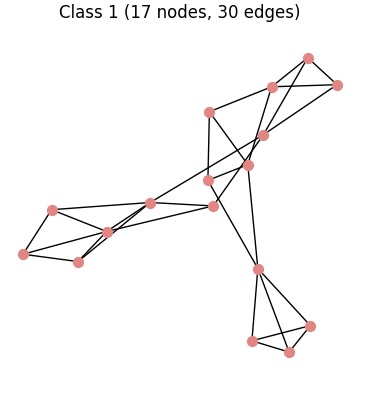}
    \end{minipage}\hfill
    \begin{minipage}{0.24\textwidth}
        \centering
        \includegraphics[width=\linewidth]{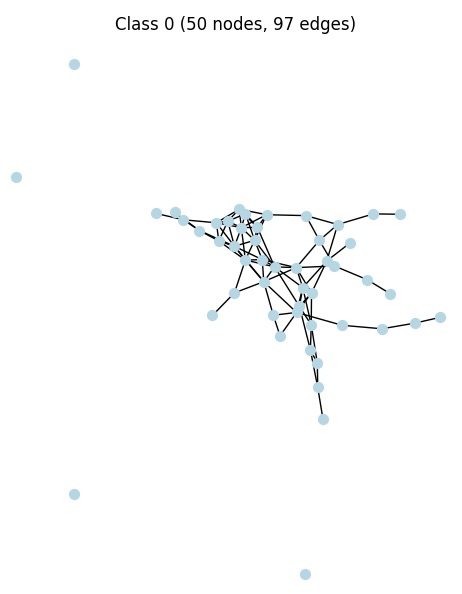}
    \end{minipage}\hfill
    \begin{minipage}{0.24\textwidth}
        \centering
        \includegraphics[width=\linewidth]{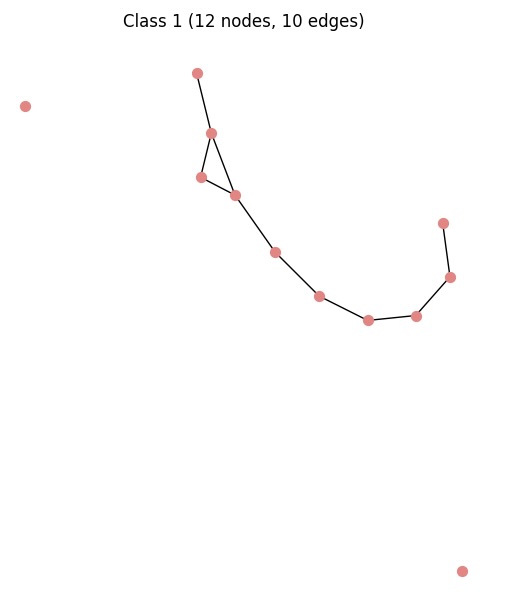}
    \end{minipage}
    \caption{Visual comparison on PROTEINS dataset (left: real graphs from Classes 0 and 1; right: generated graphs). Our model preserves structural characteristics including local motifs, triangular patterns, and community clusters through learned distance-based edge prediction. While isolated nodes appear due to deterministic density-aware selection enforcing class-average sparsity, the overall topology, clustering patterns, and class-specific structural signatures are maintained.}
    \label{fig:protein_vis}
\end{figure}

On ENZYMES, our method achieves the best clustering coefficient alignment (0.08) across all compared approaches, indicating strong capture of higher-order structural patterns. While WPGAN achieves the lowest degree MMD (0.02), our approach remains competitive (0.09) and demonstrates superior spectral performance (0.04), suggesting balanced modeling of both local and global structural properties.

Figures~\ref{fig:proteindist} and~\ref{fig:protein_vis} provide distributional and qualitative analysis on the PROTEINS dataset. The clustering coefficient histogram (Figure~\ref{fig:proteindist}, bottom) shows strong overlap between real and generated distributions, confirming superior preservation of local connectivity patterns. The degree distribution (Figure~\ref{fig:proteindist}, top), however, reveals more conservative assignment with concentration in bins 2--6 compared to the broader real distribution. Qualitatively (Figure~\ref{fig:protein_vis}), generated graphs successfully preserve key structural characteristics including triangular motifs and community-like clusters, with our distance-based edge predictor learning meaningful node relationships that produce similar topological organization. However, isolated or low-degree nodes appear more frequently in generated samples. This difference comes from our fixed top-$k$ edge selection method, which forces graphs to match class-average density but limits the natural variety in node degrees. While this ensures graphs have appropriate overall connectivity, it biases the range of individual node degrees. Despite this limitation, the model maintains overall structural consistency, clustering patterns, and class-specific characteristics while generating diverse, non-memorized instances.

A critical distinction of our evaluation is the explicit measurement of uniqueness and novelty alongside distributional metrics. Prior works~\cite{li2018learning,you2018graph,fan2019labeled,ava_cascon} focus solely on MMD scores, which can be minimized through memorization or near-replication of training samples. Our results demonstrate competitive MMD performance while maintaining high uniqueness and novelty, confirming genuine generative capability rather than overfitting. 
\end{description}

\section{Conclusion}

We propose a density-aware, class-conditional graph generation framework addressing fundamental challenges in realistic graph synthesis: discrete topology, variable graph sizes, and class-specific structural patterns. The approach learns graph connectivity in latent space through a distance-based edge predictor instead of random edge sampling with fixed probabilities, capturing meaningful relational dependencies between nodes while explicitly controlling class-specific graph density. Integrated within a Wasserstein GAN framework with gradient penalty and a GCN-based critic, the proposed model ensures stable training and promotes graphs with coherent topology and realistic connectivity patterns. 

Experiments on benchmark datasets (MUTAG, ENZYMES, and PROTEINS) demonstrate the model accurately captures statistical properties of real graphs, achieving competitive or superior performance on MMD-based distributional metrics, particularly for clustering and spectral characteristics reflecting higher-order connectivity. The framework consistently generates diverse and novel graphs, with high uniqueness and novelty scores, while preserving class-specific structural signatures and sparsity patterns. Compared to existing methods, our approach shows improved balance between structural fidelity and generative diversity, producing graphs whose local neighborhoods and global organization more closely match real-world distributions.

However, our top-$k$ edge selection, while enforcing class-specific sparsity, can biases degree distributions by reducing natural variation within classes. This suggests fixed class-average density is too restrictive for capturing degree variety in real graphs. Future work includes using probabilistic sampling from learned distributions to add variation while maintaining expected density, and adding degree distribution objectives to the training loss. 

%Appendixes go here
%\appendix

% All references should be stored in the file "references.bib".
% That call to use that file is in "cai.cls". 
% Please do not modify anything below this line.
\printbibliography[heading=subbibintoc]

\end{document}